\documentclass{article}
\usepackage[final]{corl_2018} 
\usepackage{times}
\usepackage[numbers]{natbib}
\usepackage{multicol}
\usepackage{hyperref}
\usepackage{url}
\usepackage[pdftex]{graphicx}
\usepackage{amsfonts}
\usepackage{subfig}
\usepackage{caption}
\usepackage[vlined,linesnumbered,ruled,resetcount]{algorithm2e}
\usepackage{amsmath}
\usepackage{bbm}
\usepackage{framed}
\usepackage{balance}
\usepackage{color}
\usepackage{blindtext}

\author{
  Ajinkya Jain\\
  Department of Mechanical Engineering\\
    University of Texas at Austin, USA\\
  \texttt{ajinkya@utexas.edu} \\
    \And
  Scott Niekum\\
  Department of Computer Science\\
  University of Texas at Austin, USA\\
  \texttt{sniekum@cs.utexas.edu} \\
}

\title{Efficient Hierarchical Robot Motion Planning \\ Under Uncertainty and Hybrid Dynamics}

\begin{document}
\maketitle

\vspace{-10mm}
\begin{abstract}
Noisy observations coupled with nonlinear dynamics pose one of the biggest challenges in robot motion planning. By decomposing nonlinear dynamics into a discrete set of local dynamics models, hybrid dynamics provide a natural way to model nonlinear dynamics, especially in systems with sudden discontinuities in dynamics due to factors such as contacts. We propose a hierarchical POMDP planner that develops cost-optimized motion plans for hybrid dynamics models. The hierarchical planner first develops a high-level motion plan to sequence the local dynamics models to be visited and then converts it into a detailed continuous state plan. This hierarchical planning approach results in a decomposition of the POMDP planning problem into smaller sub-parts that can be solved with significantly lower computational costs. The ability to sequence the visitation of local dynamics models also provides a powerful way to leverage the hybrid dynamics to reduce state uncertainty. We evaluate the proposed planner on a navigation task in the simulated domain and on an assembly task with a robotic manipulator, showing that our approach can solve tasks having high observation noise and nonlinear dynamics effectively with significantly lower computational costs compared to direct planning approaches.
\end{abstract}

\keywords{POMDP, Manipulation Planning, Hybrid Dynamics} 
\vspace{-2mm}
\section{Introduction}
\vspace{-2mm}
One of the biggest challenges in robot motion planning is to develop feasible motion plans for systems having highly nonlinear dynamics in the presence of partial or noisy observations. Often, these nonlinearities are caused by sudden transitions or discontinuities in the dynamics (for example, due to contacts in a robot manipulation task). When task dynamics change suddenly in state space, even small state estimation errors can lead to large deviations and plan failure. Therefore, reasoning about uncertainty over states becomes crucial in order to develop robust motion plans. Planning problems under uncertainty are often represented as a \textit{partially observable Markov decision process} (POMDP) \citep{thrun2002probabilistic}. POMDP problems have been shown in literature to be PSPACE-complete \citep{papadimitriou1987complexity}, making exact planning intractable. 

To make planning tractable, POMDP planners typically leverage various types of approximations  \citep{pineau2002policy, Brunskill2008, toussaint2008hierarchical, kurniawati2008sarsop, shani2013survey,agha2014firm, hollinger2014sampling} or structural assumptions \citep{agha2014firm, sreenath2013partially,van2012motion, indelman2015planning, majumdar2017funnel} that simplify the problem.  In this work, we propose to leverage a natural, simplifying assumption that the nonlinear dynamics of robot motion planning tasks can be decomposed into a discrete set of simpler local dynamics models, of which only one is active at any given time (e.g. a change in dynamics due to contact). Note that these local dynamics models may be approximate, especially when they are learned from data or are a simplification of a complex underlying model. A complete dynamics model can then be defined as a \textit{hybrid dynamics} model having hybrid states comprised of the continuous states of the system along with a discrete state denoting the active local dynamics model.  

The primary contribution of this work is a novel POMDP planner that plans in a hybrid belief space, allowing for efficient information gathering and planning under uncertainty with hybrid dynamics. We define a hybrid belief to represent uncertainties over the robot state and the active dynamics model and formulate hybrid belief propagation equations. Using the hybrid belief representation, a hierarchical POMDP motion planner is presented that solves the POMDP problem by dividing it into two levels: at the higher level, discrete state plans are generated to find a sequence of local models that should be visited during the task, and at the lower level, these discrete state plans are converted into cost-optimized continuous state belief-space plans.

The biggest advantage of dividing the planning problem into two levels is that it breaks long-horizon planning problems into multiple smaller segments that can be sequenced to find a complete solution. Since POMDP planning becomes exponentially more difficult with longer horizons (\citep{papadimitriou1987complexity}), a hierarchical approach breaks the problem into chunks that can be solved with significantly less effort. Another major benefit of discrete state planning is that the planner can chose to leverage a specific local dynamics model in order to improve the effectiveness of the generated plans. For example, if it is known a priori that in the $k$-th local dynamics model, motion is allowed only along a particular vector (e.g. due to presence of a wall), it can be used to reduce the state uncertainty along the dimensions orthogonal to the allowed motion vector. This indirect feedback for uncertainty reduction is critical for tasks in which observations are highly noisy, or even entirely unavailable (for example, due to occlusions).


\subsection{Related Works} 
Broadly, POMDP solving approaches can be divided into two categories based on whether their state, action and observation spaces are \textit{discrete} or \textit{continuous}. Discrete space POMDP solvers, in general, either approximate the value function using point-based methods \citep{kurniawati2008sarsop, shani2013survey} or use Monte-Carlo sampling in the belief space \citep{silver2010monte, kurniawati2016online} to make the POMDP problem tractable. Continuous space POMDP solvers often approximate the belief over states as a distribution having finite parameters (typically Gaussian) and either solve the problem analytically using gradients \citep{van2012motion, indelman2015planning, majumdar2017funnel} or use random sampling in the belief space \citep{agha2014firm, hollinger2014sampling}. Other approaches have also extended point-based methods to continuous domains \citep{shani2013survey}. 

Discrete space POMDP solvers have been shown to be able to successfully plan for large discrete space domains. However, continuous space domains are infinite-dimensional, and discrete space solvers often fail to find feasible solutions for planning horizons longer than a few steps \citep{kurniawati2016online}.
Among continuous space POMDP solvers, \citet{agha2014firm} and \citet{hollinger2014sampling} have proposed sampling based methods that can find effective solutions even in complex domains. However, most sampling based methods suffer from the problem of obtaining sub-optimal solutions and can only be probabilistically optimal at best \citep{elbanhawi2014sampling}. POMDP solvers for hybrid domains, such as the one discussed in this work, have been previously discussed by \citet{Brunskill2008},  \citet{sreenath2013partially} and \citet{agha2014health}. In the most closely related work to ours, \citet{Brunskill2008} proposed a point-based POMDP planning algorithm, the SM-POMDP planner, which approximates the nonlinear system dynamics using a multi-modal hybrid dynamics model. However, unlike our POMDP planner, the SM-POMDP planner plans only in the continuous domain and the discrete states are obtained ``passively'' using the switching conditions.

\citet{pineau2002policy} and \citet{toussaint2008hierarchical} have also previously proposed hierarchical POMDP planners. The planner developed by \citet{pineau2002policy} leverages a human-designed task hierarchy to reduce problem complexity, while \citet{toussaint2008hierarchical} emphasizes automatic discovery of hierarchy in state space using a dynamic Bayesian network. Although such approaches can work well for some robot control tasks, we believe that a more natural hierarchy of subtasks emerges automatically if a hybrid dynamics model is used to represent tasks with nonlinear dynamics, such as robot manipulation tasks involving contacts.


As hybrid dynamics models are very effective in modeling nonlinearities that are due to sudden transitions in the dynamics, a natural application for the proposed POMDP solver is contact-rich robotic manipulation. One of the current approaches for solving the robot manipulation planning problem is to search for an optimal sequence of parameterized manipulation actions or primitives to perform the task \citep{dogar2012planning, kroemer2015towards}. \citet{kroemer2015towards} have proposed to represent primitives for different phases (modes) of a multi-phase manipulation task using \textit{dynamic movement primitives (DMPs)} and learn a library of such manipulation skills which can be sequenced to perform a task. Unfortunately, a lack of a task dynamics model prevents these methods from generalizing to novel manipulation tasks, e.g. having different cost functions, even if it involves the same objects. 


\section{Background}
\textbf{POMDPs: } \label{sec:sec_3_1} Partially Observable Markov Decision Processes (POMDPs) provide a mathematical framework for the problem of sequential decision making under uncertainty \citep{thrun2002probabilistic}. Let $\mathcal{X} \subset \mathbb{R}^n$ be the space of all possible states $\textbf{x}$ of the robot, $\mathcal{U} \subset \mathbb{R}^m$ be the space of all possible control inputs $\textbf{u}$ and $\mathcal{Z} \subset \mathbb{R}^k$ be the space of all possible sensor measurements $\textbf{z}$ the robot may receive. To account for state uncertainty, a distribution of the state $\mathbf{x}_t$ of the robot given all past control inputs and sensor measurements is defined as the \textit{belief} $b[\mathbf{x}_t] \in \mathbb{B}$, given as:
\begin{equation}
    b[\mathbf{x}_t] = p[\mathbf{x}_t | \mathbf{x}_0, \mathbf{u}_0, ..., \mathbf{u}_{t-1}, \mathbf{z}_1,...,\mathbf{z}_t]
\end{equation}
where $\mathbf{x}_t \in \mathcal{X}$, $\mathbf{u}_t \in \mathcal{U}$ and $\mathbf{z}_t \in \mathcal{Z}$ are the robot's state, control input and received measurement at time step $t$, respectively and $\mathbb{B} \subset \{ \mathcal{X} \xrightarrow{} \mathbb{R} \}$ represent the space of all possible beliefs. In the general case, considering a stochastic dynamics and observation model for the process:
\begin{equation}
    \mathbf{x}_{t+1} \sim p[\mathbf{x}_{t+1} | \mathbf{x}_t, \mathbf{u}_t], ~~~~~~~ \mathbf{z}_t \sim p[\mathbf{z}_t|\mathbf{x}_t]
    \label{eq:eqn0}
\end{equation}
For a given control input $\mathbf{u}_t$ and a measurement $\mathbf{z}_{t+1}$, belief propagation using Bayesian filtering can be written as:
\begin{equation}
    b[\mathbf{x}_{t+1}] = \eta p[\mathbf{z}_{t+1} | \mathbf{x}_{t+1}] \int p[\mathbf{x}_{t+1} | \mathbf{x}_t, \mathbf{u}_t]b[\mathbf{x}_t]d\mathbf{x}_t
\end{equation}
where $\eta$ is a normalizing constant independent of $\mathbf{x}_{t+1}$.\\

\textbf{Hybrid Dynamics:}
A hybrid dynamics model of a system is a dynamics model in which the states of the system evolve with time over both continuous space $x \in X = \mathbb{R}^N$ and a finite set of discrete states $q \in Q \subset \mathbb{W}$ \citep{lygeros2012hybrid}. Each discrete state of the system corresponds to a separate dynamics model that governs the evolution of continuous states. These types of dynamical models are sometimes referred to as \textit{switched dynamical systems} in the literature \citep{ghahramani2000variational}.  

In a hybrid model, discrete state transitions of the system can be represented as a directed graph with each possible discrete state $q$ corresponding to a node and edges ($e \in E \subseteq Q \times Q$) marking possible transitions between the nodes. These transitions are conditioned on the continuous states. A transition from the discrete state $q$ to another state $q'$ happens if the continuous states $\mathbf{x}$ are in the \textit{guard set} $G(q, q')$ of the edge $e^{q'}_{q}$ where $e^{q'}_{q} = \{q, q'\}$, $G(\cdot): E \to P(X)$ and $P(X)$ is the power set of $X$. Thus, for each discrete state $q$, in a hybrid dynamics model we can define:
\begin{align}
\begin{split}
x_{t+1} =F^{q}(x_t,u_t), ~~~~~~~~~~~~~~
z_t = H^{q}(x_t)
\end{split}
\label{eq:eqn1}
\end{align}
where $x \in \mathbb{R}^n$, $u \in \mathbb{R}^m$, $z \in \mathbb{R}^l$, $F^q(x,u)$ and $H^q(x)$ are the continuous state, control input, observation variables, state dynamics and observation functions respectively. Evolution of the discrete state of the system can be modeled by a finite state Markov chain. Defining the state transition matrix as $\Pi = \{\pi_{ij} \}$, the discrete state evolution can be given as:
\begin{equation}
q_{t+1} = \Pi q_t
\label{eq:eqn2}
\end{equation}

\section{Hierarchical POMDP Planner}
\label{sec:sec4}
We propose to solve the problem of motion planning under uncertainty for tasks governed by highly nonlinear dynamics as a POMDP problem defined on a hybrid dynamics model. Different local dynamics models constituting the task dynamics are represented as distinct discrete states of the hybrid model. Under uncertainty over the robot state, a separate discrete distribution needs to be maintained to represent our confidence over the active local dynamics model at each time step. Jointly, a hybrid belief over the hybrid state of the system can be defined with a continuous part representing uncertainty over the robot state and a discrete part representing uncertainty in the active local dynamics model. In this work, we assume that the continuous part of hybrid belief is represented by a mixture of $\mathcal{L}$ Gaussian distributions, each having a mixing weight of $\alpha_l$, given as:
\vspace{-5pt}
\begin{equation}
    b^x_t = \sum^L_{l=1} \alpha_l \mathcal{N}(\mu_l, \Sigma_l)
\end{equation}

\subsection{Belief Propagation under Hybrid Dynamics}
A hybrid belief is defined as $B = \{b^{x}, b^{q}\}$, where $b^{x}$ and $b^{q}$ correspond to the belief over continuous robot state, $\mathbf{x}$, and discrete states, $q$, respectively. Propagation of hybrid beliefs using Bayesian filtering can be separated into two steps: making a prediction using the dynamics model to obtain a belief \textit{prior} and updating it based on the received observation to compute the belief \textit{posterior}. 

\subsubsection{Belief Prior}\vspace{-5pt}
We extend the system dynamics, $F^{q}(x_t, u_t)$, for uncertainty propagation and represent it as $\mathcal{F}^{q}(b^x_{t}, u_t)$. At each time step $t$, we can propagate the current belief $b^x_t$ through the system dynamics of each discrete state $\mathcal{F}^{q}(x_t, u_t)$ individually and then take a weighted sum of the propagated belief set to obtain a belief prior for the next time step $\hat{b}^x_{t+1}$, as:
\begin{equation}
\hat{b}^x_{t+1} = \sum_{q'} \mathcal{F}^{q'}(b^x_{t}, u_t) ~b^{q}_t[q']
\label{eq:eqn5}
\end{equation}
where $b^{q}_t[q'] = p(q_t=q'|x_t)$ is $q'$-th component of $b^q_t$, and $x_t$, $q_t$ and $u_t$ represent the continuous states, discrete state, and continuous control input to the system at time $t$, and $\hat{b}[\mathbf{x}_{t+1}]$ is denoted as $\hat{b}^x_{{t+1}}$. Note that $\mathcal{F}^{q'}(x_t, u_t)$ represents a general dynamics function and can be stochastic. Under stochastic continuous state dynamics, the definition of the discrete state transition matrix as given in Equation~\ref{eq:eqn2} needs to be extended. Assuming the transitions of discrete states are given by a directed graph with self-loops, we can define the extended discrete state transition matrix $\Pi$ at time $t$ as $\Pi_t = \{p(q^j_{t+1} | q^i_{t}, \hat{b}_{t+1}^x)\} ~~\forall q^i, q^j \in Q$ where 
\begin{equation}
p(q^j_{t+1} | q^i_{t}, \hat{b}_{t+1}^x)  = 
  \begin{cases}
                ~\eta \int_{\mathbb{R}^N}  \mathbbm{1}^{ q^j}_{q^i} (\mathbf{x}) \hat{b}^x_{t+1}(\mathbf{x}) d\mathbf{x}, ~~~~~~~~~~~if~\exists ~e^{q^i}_{q^j},\\
                ~~\epsilon, ~~~~~~~~~~~~~~~~~~~~~~~~~~~~~~~~~~~~~~~~~~~~~~~ otherwise
    \end{cases}
\label{eq:eqn4}    
\end{equation}

\vspace{-2mm} where $\mathbbm{1}^{ q^j}_{q^i}(\mathbf{x})$ is an indicator function defined as:
\begin{equation}
\mathbbm{1}^{ q^j}_{q^i}(\mathbf{x}) =
    \begin{cases}
            1, & ~~~~~~\text{if}\ \mathbf{x} \in G(q^i, q^j)\\
            0, & ~~~~~~otherwise
    \end{cases}
\end{equation}
where $\eta = \sum^{|Q|}_{k=1} \pi_t(i, k)$ is a normalization constant, and $\epsilon$ is a small probability to handle cases when received observations do not correspond to any legal discrete transition. Calculating the extended discrete state transition matrix $\Pi_t$ at each time step  using Eq.~\ref{eq:eqn4} can be computationally expensive. An approximation of $\Pi_t$ can be obtained by sampling $n$ random points from the belief over continuous states $b^x_{t+1}$ and calculating ratio of points lying in the guard set $G(q^i, q^j)$ to the total number of sampled points for each discrete state $q^j$.

\subsubsection{Belief Posterior}
We use a hybrid estimation algorithm based on Bayesian filtering to reduce the uncertainty over states using noisy continuous state observations. The proposed algorithm consists of two layers of filters: the first estimates the continuous states of the system and the second estimates the discrete states of the system. Upon receiving observation $z_{t+1}$, the continuous state prior is updated by taking a weighted sum of a bank of extended Kalman filters running independently, with each discrete mode having an individual filter. The weights for the sum are determined using the prior for the discrete mode ${\hat{b}^q_{t+1}}$. The complete update step for continuous states can be written as:
\begin{equation}
b^x_{t+1} = \hat{b}^x_{t+1} + \sum_{q'} \Big( \mathbf{K}^{q'}_{t+1} (z_{t+1}-\mathcal{H}^{q'}_{t+1}(\hat{b}^x_{t+1}))\Big) \hat{b}^{q}_{t+1}[q']
\label{eq:eqn6}
\end{equation}
where $\mathbf{K}^{q'}_{t+1}$ is the Kalman Gain for discrete state $q'$ at time $t+1$ and $\hat{b}^{q}_{t+1}[q']$ is $q'$-th component of $\hat{b}^q_{t+1}$. The update for the discrete state can be obtained by using a Bayesian filter update given as:
\begin{equation}
b^q_{t+1} = \gamma \mathbf{M}_{t+1} \circ \hat{b}^q_{t+1}
\label{eq:eqn7}
\end{equation}
where $\mathbf{M}_{t+1} = [P(z_{t+1}|q_{t+1}=q')]^T ~~\forall q' \in Q$, $\circ$ is the element-wise multiplication operator, $\gamma = \dfrac{1}{\sum_{q'} \mathbf{M}_{t+1} \circ \hat{b}^{q'}_{t+1}}$ is a normalization constant and
\begin{equation}
P(z_{t+1}|q_{t+1}=q') = z_{t+1} \sim \mathcal{H}^{q'}_{t+1}(b^x_{t+1})
\end{equation}
where $\mathcal{H}^{q'}_{t+1}(.)$ is the observation function for state $q'$. Mixing weights for the mixture of Gaussians are also updated based on the received observations as
\begin{equation}
\alpha^l_{t+1} = \mathcal{N}(z_{t+1} - \hat{z}^l_{t+1}| \mathbf{0},\Sigma^l_{t+1}), ~~~~where,~~~ \hat{z}^l_{t+1} = \sum_{q'} \hat{b}^{q}_{t+1} [q'~]~(\mathcal{H}^{q'}_{t+1} \mu^l_{t+1})
\end{equation}
A new mixture of $\mathcal{L}$ Gaussians is then chosen to represent the continuous belief $b^x_{t+1}$ at next step. 

\vspace{-5pt}
\subsection{Direct Planning}
With the hybrid belief propagation equations defined, we can now use trajectory optimization technique to solve the POMDP. We assume \textit{maximum likely observations (MLO)} obtained by propagating the current belief over continuous states through the system dynamics (Eqn.~\ref{eq:eqn5}) as the true observations for developing locally optimal motion plans, as introduced by Platt et al. \citep{Jr2010}. In this work, the nonlinear optimization problem set up for trajectory optimization is posed as a sequential least squares programming (SQP) problem and solved using the SNOPT software package \citep{GilMS05, snopt76}. We denote this approach as the \textit{direct planning} approach. 

\begin{algorithm}[t]
  \footnotesize
\SetKwProg{Fn}{Function}{}{}
\Fn{$high\_level\_plan\_to\_countinuous\_state\_goals$ (high-level plan)}{
    \For{each $q^{k}$ in high-level plan}{
        Define corresponding full-confidence vector,  $W^{k}_{full\_conf}$ = 
            $\begin{cases}
                1, & \text{if}\ q = q_{goal}^{k} \\
                0, & \text{else} 
            \end{cases}$\\
        Sample $n$ random points:  $X_{sample}$ = $\{x_1,...,x_n \} \sim \mathcal{X}$; \\
        \For{each $x_i \in X_{sample}$}
        {
            Find confidence distribution on discrete states $w_i \in W_{sample}$: \\
            \hspace{0.3cm}{Sample a random set $X' \sim \mathcal{X}$; \\
            \hspace{0.3cm}\For{each $q' \in \mathcal{Q}$}{
                \hspace{0.2cm}$w_i(q') = \dfrac{|x' \in X' \cap G(q', q'')~\forall q''|}{|X'|}$ ; \\
                }}
            Find cost of divergence $c_i \in C' \subset \mathbb{R}$:
            $c_i(x_i)$ = $Hellinger(w_i, W^{k}_{full\_conf})$;\\
        }
        Define cost map on complete domain $\mathcal{X}$:
        $C_{complete}$(\textbf{x}) = Interpolate($C'$); \\
        Find best representative point in continuous state:
        $x^k_{best}$ = $global\_optimization$(\textbf{x}, $C_{complete}$); \\
        Append $x^k_{best}$ to $X_{cs\_goals}$;
        }
    \textbf{return} $X_{cs\_goals}$;
    }
\caption{ High-Level Plan $\boldsymbol{\xrightarrow{}}$ Continuous State Goals}
\label{alg:alg1}
\end{algorithm}

\vspace{-3pt}
\subsection{Hierarchical Planner}
Although the direct planning approach can be used to solve the POMDP, planning for longer horizons in complex tasks, such as contact-rich manipulation tasks, can result in infeasible computational costs \citep{papadimitriou1987complexity}. To tackle this challenge, we propose a hierarchical planner that decomposes the POMDP problem into smaller subproblems which can be solved with significantly less effort. 

The proposed hierarchical planner has two levels: a higher level to find the best sequence of local dynamics models that should be visited along the path (by visiting corresponding regions in continuous state space) and a lower level that is similar to the aforementioned direct planning approach. The higher level planner generates a set of candidate high-level plans consisting of all feasible permutations (without repetitions) of the discrete states of the task\footnotemark[1]. A transition between two discrete states is deemed to be an infeasible transition, if the regions of the continuous state space corresponding to the two discrete states form a pair of positively-separated sets.


\footnotetext[1]{The feasibility check also helps in keeping the POMDP tractable. \citet{gulyas2011estimation} have shown that the average path length for a connected graph decreases as its graph connectivity increases. If the graph of discrete states, from which the set of feasible high-level plans is derived, is not sparse enough to solve the POMDP tractably, a simple heuristic can be defined that penalizes plans with longer path lengths. Preferential choice of shorter plans results in fewer calls to the lower level planner and reduces computational time.}

We define the term \textit{confidence} to denote the probability of a continuous state belief to be in a particular discrete state. Spatial distribution of confidence across the continuous domain for a particular discrete state is defined as the \textit{confidence map} associated with that state. A confidence map for a particular discrete state can be converted into a cost map by calculating a cost of divergence between a full-confidence vector ($W^{k}_{full\_conf}$, one-hot vector with probability of being in that particular state equals to one) and the confidences at randomly sampled points across the domain. A high-level plan can then be converted into a sequence of continuous state goals by finding the global minimum of such cost maps associated with each discrete state in the plan (see Algorithm~[\ref{alg:alg1}]). The lower level planner is then called for each of these continuous state goals and a complete continuous state path for the high-level plan is generated by combining the outputs of lower level planner. An additional discrete state is added to each high-level plan which represents the desired goal of the task and is considered to be active within an $\epsilon-$neighbourhood of the actual task goal. High-level plans are then ranked by calculating a divergence cost on the distribution of planner's confidence on the active discrete state at the final point of the plan and the desired confidence distribution (all the probability mass within the $\epsilon-$neighbourhood of the goal). The continuous state plan corresponding to the high-level plan with the minimum cost is chosen to be executed. 

In this work, we have used \textit{Hellinger distance} to calculate the divergence cost \citep{cha2007comprehensive} between the discrete distributions as it forms a symmetric bounded metric with a value between 0 and 1, and was found to be more numerically stable than the Bhattacharya distance, KL-divergence, and Jensen–Shannon divergence on the tested application domains. Radial basis functions were used to interpolate the divergence costs throughout the domain and the differential evolution method was used to find the approximately globally optimal solutions of the generated cost map \citep{storn1997differential}.

\vspace{-3pt}
\subsection{Trajectory Stabilization}
With the MLO assumption, it is very likely that during execution the belief over robot state will diverge from the \textit{nominal} trajectory planned. To ensure that the execution phase belief follows the plan, a \textit{belief space LQR (B-LQR)} controller can be defined around the nominal trajectory. B-LQR controllers were introduced by Platt et. al \citep{Jr2010} and can be seen as belief-space extension of Linear-Quadratic Regulators (LQR). For systems modelled as linear-Gaussian processes, a B-LQR controller is optimal and equivalent to a linear-Quadratic Gaussian (LQG) controller. In B-LQR, each point in the nominal trajectory is defined as a set point and quadratic costs are defined for the distance from it and the control effort required to converge to it. Closed form solutions exist to ensure convergence to the set point within a finite time horizon. While stabilizing the trajectory, the most likely active discrete state is taken to define the governing dynamics of the system. However, it may happen that the B-LQR controller is unable to stabilize the execution phase (actual) belief around the nominal trajectory. If the planned belief for the next step deviates more than a $\delta$-threshold from the actual belief after the observation update, a replanning call to the planner is triggered.

\section{Experiments}

\subsection{Domain-I: Walled Domain}
The first task is an autonomous navigation task in a 2D domain ($\{x,y\} \in [-2,15]$) having extremely noisy observations (zero-mean Gaussian noise $w \sim \mathcal{N}(\cdot|0, 15~units)$). The domain consists of two perpendicular walls parallel to the x and y axis respectively. As the motion along a wall is constrained to be only parallel to the wall, the robot can use it to efficiently localize itself in a direction orthogonal to the wall. We compare the performance of the hierarchical planner with the direct planning approach. Note that the direct planning approach is similar in principle to the SM-POMDP planner proposed by \citet{Brunskill2008} and hence, provides a comparison of the proposed hierarchical planner with a flat, single-level planning approach. Hybrid dynamics model can be given as
{\footnotesize
\begin{align*}
    f(\mathbf{x}_t,\mathbf{u}) = 
    \begin{cases}
    \mathbf{x}_t + \mathbf{u}, & \text{if}\ x>-2, y>-2\\
    \mathbf{x}_t + \left[ {\begin{array}{cc}
      		             0 & 0 \\ 
                         0 & 1 \end{array} } \right]\mathbf{u}, & \text{if}\ x<-2\\ 
    \mathbf{x}_t + \left[ {\begin{array}{cc}
      		             1 & 0 \\ 
                         0 & 0 \end{array} } \right]\mathbf{u}, & \text{if}\ x>-2, y<-2,\\
    \end{cases}
\end{align*}}
where $\mathbf{x}_t = \{x_t, y_t\}^T$. The observation function was defined as $h(x_t) = x_t + w$. 

\begin{figure}[t]%
    \centering
    \subfloat[Direct Planning]{{\includegraphics[height=0.32\linewidth, width=0.35\linewidth]{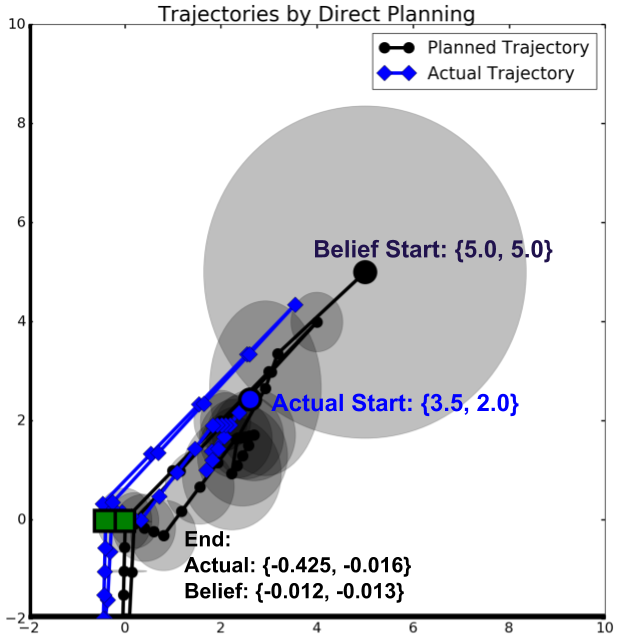} }}%
    \qquad
    \subfloat[Hierarchical Planning]{{\includegraphics[height=0.32\linewidth, width=0.35\linewidth]{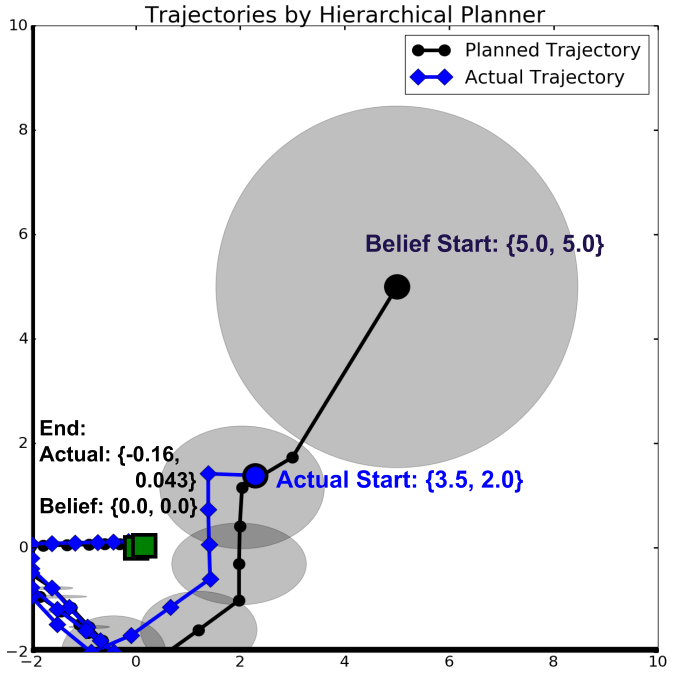}}}%
    \caption{A comparison of planned and actual trajectories using the direct planning and hierarchical planning approaches on the walled domain. For both cases, Initial belief mean $\mu =\{5,5\}$, $cov = diag(11.5, 11.5)$ , True start position:=\{3.5, 2.0\}. Gray circles represent belief covariance.}%
    \label{fig:fig2}
\end{figure}

Sample trajectories planned by the direct planning and the hierarchical planner are shown in Figure~\ref{fig:fig2}. It is evident from the figures that the hierarchical planner plans to selectively visit the two discrete states representing the walls, in contrast to the direct method. Also, the hierarchical planner is able to converge to the goal faster and with a much lower uncertainty than the direct planning approach. As the direct planner does not leverage the knowledge of local dynamics models in a structured way, it needs to plan longer trajectories to gather more information. However, due to high noise in the observations, it still fails to converge to the goal with high accuracy. 

Additional statistical analysis to compare the two approaches in terms of total planning time, final error and final belief uncertainty are presented in Table~\ref{table:table1}. It can be seen from Table~\ref{table:table1} that, for comparable final error and final belief uncertainty, the hierarchical planner is able to find a solution approximately $5$ times faster than the direct planning approach. 

\begin{table}[b]
    \centering
    \begin{tabular}{ @{}| l | l | l |@{}}
    \hline
    \textbf{Metric} & \textbf{Direct } & \textbf{Hierarchical } \\ \hline
    Average Total time (in seconds) & 51.908 & 10.695 \\ \hline
    Average Final Error & $[-0.168, 0.172]^T$ & $[0.086, 0.198]^T$ \\ \hline
    Average Final Maximum Belief Uncertainty & 0.696 & 0.625 \\
    \hline
    \end{tabular}
    \vspace{3pt}
    \caption{Comparison of direct and hierarchical planning. Values are averaged over 5 runs. Planning horizon: 20 steps. Belief start: $[5,5]^T$. actual start: $[3.5,2.0]^T$. Termination condition: Maximum likelihood estimate of belief converged within a ball of $0.2~unit$ radius around the goal ($[0,0]^T$) with max covariance of $1~unit$.}
    \label{table:table1}
\end{table}

\subsection{Domain-II: Airplane assembly}
\vspace{-2pt}
We experimentally demonstrate that the hierarchical POMDP planner can be used to tractably solve a real world manipulation task --- the partial assembly of a toy airplane from the YCB dataset \citep{calli2017yale}. We considered the first step of inserting the landing gear into the wing as a test case for our planner. The task requires a high precision with maximum tolerance of $\pm 0.2~cm$. Feedback on the location of the airplane in world was noisy and had an average estimation error of $\pm 2.0~cm$. This experiment demonstrates two important features of the proposed planner: first, the planner can be scaled to solve real-world manipulation planning under uncertainty problems and second, due to the hierarchical planning approach, the planner essentially enables the robot to plan and ``feel around" to localize itself when observations are noisy, similar to what a human might do. 

In a robot manipulation task involving contacts, based on the type of contact between the bodies, the number of state-dependent local dynamics models can be large, or even infinite. We simplify the problem by assuming an approximate hybrid dynamics model, in which the local dynamics models correspond to possible motion constraints that the robot can encounter while executing the task. For example, the task of placing a cup on a table can be considered to be approximately made of two local dynamics models: one when the two objects are not in contact and the other when the cup is in contact with the table plane. The second dynamics model represents the motion constraint placed on the cup by the table by restricting its motion to be only along its plane and not penetrating it. This approximation helps in having a succinct and effective representation of the task dynamics; under this approximation, for a specific set of inputs, the relative motion between the two objects in contact will always be the same, independent of the type of contact between them. In this case, the specific set of inputs would be the set of all inputs which do not result in moving the cup away from the table plane, resulting in breaking the contact between them. 

\begin{figure}[t]
    \centering
    \subfloat{{\includegraphics[height=3.5cm, width=8.cm]{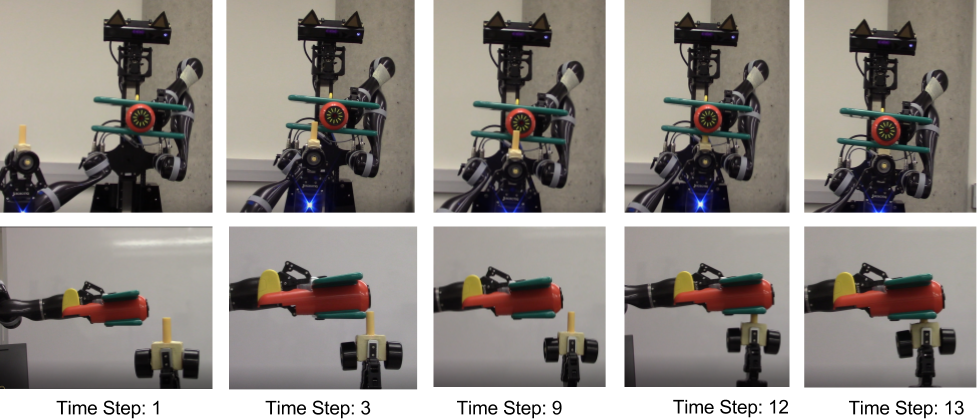} }}%
    \qquad
    \subfloat{{\includegraphics[width=5.cm,height=4.cm]{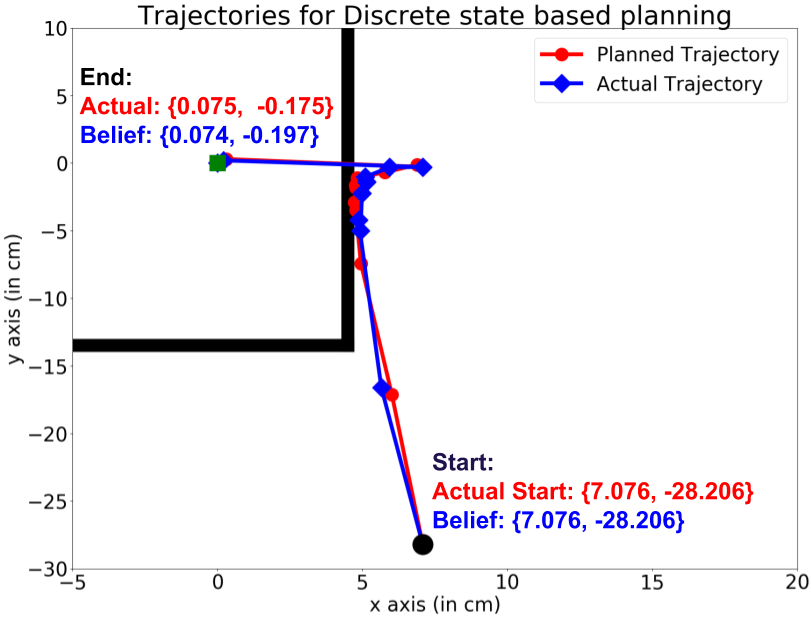} }}%
    \caption{\textit{Left Panel}: Snapshots of the robot assembling the toy airplane. \textit{Right Panel}:Planned and Actual trajectories for the airplane assembly task in a plane parallel to wing plane. Bold black lines represents the edges of the airplane wing. 1 unit = 10 cm.}
    \label{fig:fig7}%
\end{figure}

In this experiment, we consider the domain to be made up of four distinct local dynamics models: two corresponding to the linear motions along the wing plane edges, one corresponding to the corner of the plane and one to represent free-body motion elsewhere in the domain. At the highest level, the planning problem can be broken down into two steps: first, to localize the gear at a point in a plane parallel to the wing and second, to insert the gear into the hole. A hybrid dynamics model in a plane parallel to the wing can be given as
{\footnotesize
\begin{align}
    f(\mathbf{x}_t,\mathbf{u}) = 
    \begin{cases}
        \mathbf{x}_t + \left[ {\begin{array}{cc}
      		             0 & 0 \\ 
                         0 & 1 \end{array} } \right]\mathbf{u} + \mathbf{v}, & \text{if}\ x\in[4, 4.5], y>-13.5\\ 
        \mathbf{x}_t + \left[ {\begin{array}{cc}
      		             1 & 0 \\ 
                         0 & 0 \end{array} } \right]\mathbf{u} + \mathbf{v}, & \text{if}\ x<4, y \in [-14, -13]\\
        \mathbf{x}_t + \mathbf{0}*\mathbf{u} +  \mathbf{v}, & \text{if}\ x \in [4, 4.5], y \in [-14, -13.5]\\
        \mathbf{x}_t + \mathbf{u} + \mathbf{v}, & \text{otherwise}\\
    \end{cases}
\end{align}}

where $\mathbf{v}$ is process noise, modeled as $v \sim \mathcal{N}(\cdot|0, \mathbf{I}_2)$ with 1 unit = 1 $cm$. The observation function $h(x_t) = x_t + w$ with zero-mean Gaussian observation noise $w \sim \mathcal{N}(\cdot|0, 2\mathbf{I}_2)$. The planner took $14.682$ seconds for planning on an Intel\textsuperscript{\textregistered} Core\textsuperscript{TM} i7-6700 CPU @3.40GHz, 16Gb RAM. The left panel of Figure~\ref{fig:fig7} shows snapshots of the trajectory executed by the robot during the task from two perpendicular angles. The right Panel of Figure~\ref{fig:fig7} shows the trajectory planned by the hierarchical planner and the actual trajectory taken by the robot in a plane parallel to the wing. It can be see from Fig.~\ref{fig:fig7} that the planner plans to activate the motion constraint parallel to the wing in order to reduce its uncertainty. Once localized in the plane parallel to the wing, the robot changes planes to move to a point directly above the hole and then proceeds to insert the landing gear into the wing.

\section{Conclusion}
\vspace{-3pt}
Nonlinear task dynamics, especially due to sudden changes in dynamics, can be effectively modelled using a hybrid dynamics model. A hybrid dynamics model consists of a set of local dynamics models, of which only one is active at a time. In this work, we propose a hierarchical POMDP planner for hybrid dynamics which can develop locally optimal motion plans for tasks involving nonlinear dynamics under noisy observations. The proposed planner generates hierarchical motion plans at two levels: first, a high-level motion plan that sequences the local dynamics models to be visited and second, based on the best high-level plan, a detailed continuous state motion plan to be followed by the robot. The hierarchical planning approach breaks the large POMDP problem into multiple smaller segments with shorter planning horizons, which significantly increases the computational efficiency of the planner. High-level planning also enables the robot to leverage task dynamics to improve its performance---for example, reducing uncertainty using the task motion constraints in order to develop motion plans which are more robust to state uncertainty.

In the present work, a hybrid model of task dynamics needs to be provided to the planner by an expert. Hence, a natural extension of this work is to autonomously learn a task hybrid dynamics model. For example, Niekum et al. have proposed methods \citep{niekum2015online, niekum2015ActiveArtic} to learn articulated motion models encountered while manipulating an object. In the future, the proposed POMDP planner may be combined with these methods to develop an end-to-end approach for learning hybrid dynamics models for manipulation tasks and using them to generate plans that are robust to state uncertainty.

\clearpage
\acknowledgments{This work has taken place in the Personal Autonomous Robotics
Lab (PeARL) at The University of Texas at Austin. PeARL research
is  supported  in  part  by  the  NSF (IIS-1724157,  IIS-1638107,  IIS-1617639,  IIS-1749204) .}

\bibliography{references}
\clearpage
\appendix
\section{Preliminaries}
\subsection{Trajectory Optimization using Direct Transcription}
Direct Transcription is a trajectory optimization method in which a constrained nonlinear optimization problem is set up with the user-defined objective function over a set of \textit{knot-points} $\{\mathbf{x}_i$, $\mathbf{u}_i \}$) chosen to discretize the continuous space trajectory into a set of decision variables. The system dynamics are imposed as the constraints on the optimization problem. For discrete-time systems, these knot-points can be taken as the system state $\mathbf{x}_t$ and the control input $\mathbf{u}_t$ at each time step $t$. However, planning for longer horizons will then require specifying a high number of \textit{knot-points} ($\mathbf{x}_i$, $\mathbf{u}_i$) which can result in very high computational costs. This can be resolved by  approximately parameterizing the space of possible trajectories by a series of $M$ segments and solving the optimization problem for a knot points only at the start and end points of segments. The intermediate points on the trajectory can be obtained via numerical integration.  Let $x'_{1:M}$ and $u'_{1:M-1}$ be sets of state and action variables that parameterize the trajectory in terms of segments. The $i^{th}$ segment can be assumed to start at time $i\delta$ and ends at time $i\delta +\delta -1$, where $\delta = \dfrac{T}{M}$ for a time horizon $T$.

A general objective function for trajectory optimization can be given as 
\begin{equation}
J(x_{1:T}, u_{1:T}) \approx \hat{J}(x'_{1:M},u'_{1:M}) = \sum^M_{j=1}x'^T_i Q x'_i + u'^T_i R u'_i
\end{equation}

where $Q$ and $R$ represent the cost matrices associated with the state and the input respectively. The system dynamics incorporated as constraints can be defined as:
\begin{equation}
x'_{2} = \phi(x'_{1},u'_1), ~~~...~~~ x'_{k} = \phi(x'_{k-1},u'_{k-1})
\end{equation}
where the function $\phi(x'_i,u'_i)$ can be seen as performing numerical integration of the current state variable $x'_i$ till the next state variable $x'_{i+1}$. The function $\phi$ is given as 
\begin{equation}
x'_{i+1} = \phi(x'_i, u_i) = F(x'_i,u_i)+ \sum^{i \delta + \delta-1}_{t=i \delta} \left[ F(x_{t+1},u_i) - F(x_t,u_i) \right]
\end{equation}
where $F(x_t, u_t)$ represents the system dynamics.

Trajectory optimization using direct transcription can be extended for belief space planning by assuming Gaussian noise over continuous states \cite{Jr2010}. If the belief over continuous states is defined as $b_t = \mathcal{N}(\mu_t, \Sigma_t)$, trajectory optimization can be formulated as an optimization problem over variables $\mu_t$ and $s_t$, where $\mu_t$ represents the mean of the belief state and $s_t = \{s_1^T, ..., s_d^T\}^T$ is a vector composed of $d$ columns of $\Sigma_t = [s_1,..., s_d]$. Analogous to the deterministic case, problem is constrained to follow belief space dynamics. The corresponding objective function can be given as 
\begin{equation}
\begin{split}
J(b_{1:T}, u_{1:T}) &\approx \hat{J}(b'_{1:M},u'_{1:M})  \\
&= s^T_M \Lambda s_M + \sum^M_{j=1}\mu'^T_i Q \mu'_i + u'^T_i R u'_i
\end{split}
\end{equation}

where $Q$, $R$ and $\Lambda$ represent the cost matrices associated with belief mean, control input and the belief covariance at final discrete time step respectively. Belief dynamics can be incorporated in the formulation as the constraints:
\begin{equation}
    b'_{2} = \Phi(b'_{1},u'_1), ~~~...~~~ b'_{k} = \Phi(b'_{k-1},u'_{k-1})
\end{equation}
where the function $\Phi(b'_i, u_i)$ is given as
\begin{equation}
b'_{i+1} = \Phi(b'_i, u_i) = \mathcal{F}(b'_i,u_i)+ \sum^{i \delta + \delta-1}_{t=i \delta} \left[ \mathcal{F}(b_{t+1},u_i) - \mathcal{F}(b_t,u_i) \right]
\end{equation}
where $\mathcal{F}(b'_i, u_i)$ represents extended system dynamics. Propagation of belief $b_t$ through system dynamics $\mathcal{F}(b'_i, u_i)$ has been previously discussed by Platt et al. \cite{Jr2010} in further details. 

\section{Further Experimental Details}
\subsection{Domain-I}
Matrices defining the cost function over error in states, control input, additional cost for final state error and covariance were taken as $Q=diag(0.5,0.5)$, $R=diag(10.0, 10.0)$, $Q_{T} = 1e4$ and $\Lambda = 1e7$ respectively. Number of Gaussians used to model continuous belief $L=1$.

\subsection{Domain-II}
Feedback was obtained on the location of the airplane in the world frame by doing an online color-based object cluster extraction, using multi-plane segmentation from the Point Cloud Library (PCL) on the point cloud data of a Microsoft Kinect v2 sensor. Matrices defining the cost function over error in states, control input, additional cost for final state error and covariance were taken as $Q=diag(0.5,0.5)$, $R=diag(0.1, 0.1)$, $Q_{T} = 5000$ and $\Lambda = 1e7$ respectively. Number of Gaussians used to model continuous belief $L=1$.


\end{document}